%% file: arxiv.tex
\pdfoutput=1
\documentclass[11pt]{article}

\usepackage[]{acl}

\usepackage{times}
\usepackage{latexsym}
\usepackage{lipsum} 
\usepackage{graphicx}
\usepackage{graphicx}
\usepackage{graphicx}
\usepackage{subfigure}
\usepackage{amsmath}
\usepackage{amssymb}

\usepackage{latexsym}
\usepackage{graphicx}

\usepackage{graphicx}
\usepackage{amsmath}
\usepackage{amsthm}
\usepackage{booktabs}
\usepackage{algorithm}
\usepackage{algorithmic}
\usepackage{multirow}
\usepackage{amsmath}
\usepackage{amssymb}
\usepackage{graphicx}
\usepackage{graphicx}
\usepackage{subfigure}
\usepackage{times}
\usepackage{latexsym}
\usepackage{algorithmic}
\usepackage{graphicx}
\usepackage{graphicx}
\usepackage{subfigure}
\usepackage{amsmath}
\usepackage{amssymb}
\usepackage{algorithm}
\usepackage{algorithmic}

\usepackage{graphicx}
\usepackage{amsmath}
\usepackage{amsthm}
\usepackage{booktabs}
\usepackage{algorithm}
\usepackage{algorithmic}
\usepackage{multirow}
\usepackage{amsmath}
\usepackage{amssymb}
\usepackage{graphicx}
\usepackage{graphicx}
\usepackage{subfigure}

\usepackage{graphicx}
\usepackage{graphicx}
\usepackage{subfigure}
\usepackage{amsmath}
\usepackage{amssymb}
\usepackage{algorithm}
\usepackage{algorithmic}
\usepackage{latexsym}
\usepackage{lipsum}

\usepackage{graphicx}
\usepackage{amsmath}
\usepackage{amsthm}
\usepackage{booktabs}
\usepackage{algorithm}
\usepackage{algorithmic}
\usepackage{multirow}
\usepackage{amsmath}
\usepackage{amssymb}
\usepackage{graphicx}
\usepackage{graphicx}
\usepackage{subfigure}
\usepackage{lipsum}
\usepackage{wasysym}
\usepackage{bm}
\usepackage{algorithm}

\usepackage{microtype}
\usepackage{tabularx}

\usepackage{amsmath}

\usepackage[T1]{fontenc}
\usepackage[utf8]{inputenc}

\usepackage{microtype}

\usepackage{inconsolata}

\usepackage{graphicx}

\title{Textualized Agent-Style Reasoning for Complex Tasks by Multiple Round LLM Generation}

\author{Chen Liang\textsuperscript{2,3}\thanks{\ \ Equal Contribution}, Zhifan Feng \textsuperscript{1,3}\footnotemark[1], Zihe Liu\textsuperscript{2}, Wenbin Jiang\textsuperscript{4}, \\ \textbf{Jinan Xu}\textsuperscript{2}\thanks{\ \ Corresponding author.}, \textbf{Yufeng Chen}\textsuperscript{2}, \textbf{Yong Wang}\textsuperscript{1}\footnotemark[2] \\
\textsuperscript{1}	University of Science and Technology of China, Hefei, China \\
\textsuperscript{2} Beijing Jiaotong University, Beijing, China \\ 
\textsuperscript{3}	Baidu Inc, Beijing, China \\
\textsuperscript{4}	School of Artificial Intelligence, Beijing Normal University, Beijing, China \\
	\tt	\{21120367\}@bjtu.edu.cn
}

\begin{document}
\maketitle
\begin{abstract}
Chain-of-thought prompting significantly boosts the reasoning ability of large language models but still faces three issues: hallucination problem, restricted interpretability, and uncontrollable generation.
To address these challenges, we present AgentCOT, a llm-based autonomous agent framework, which can solve complex problems in an agent-style manner by multiple round LLM generation.
At each step, AgentCOT selects an action and executes it to yield an intermediate result with supporting evidence.
% until the problem is solved.
In addition, we 
% propose an enhanced self-consistency strategy to ensure the quality of response at each step and 
integrate the step‘s index into the reasoning process to form a graph structure for complex inference logic.
We introduce two new strategies to enhance the performance of AgentCOT.
% 实验
We conduct extensive experiments to verify the effectiveness of our method on six common benchmarks. Results exhibit that our method brings in substantial improvements over current competitive approaches. 
\end{abstract}

\section{Introduction}

\begin{figure}[ht]
\centering
\includegraphics[scale=0.25]{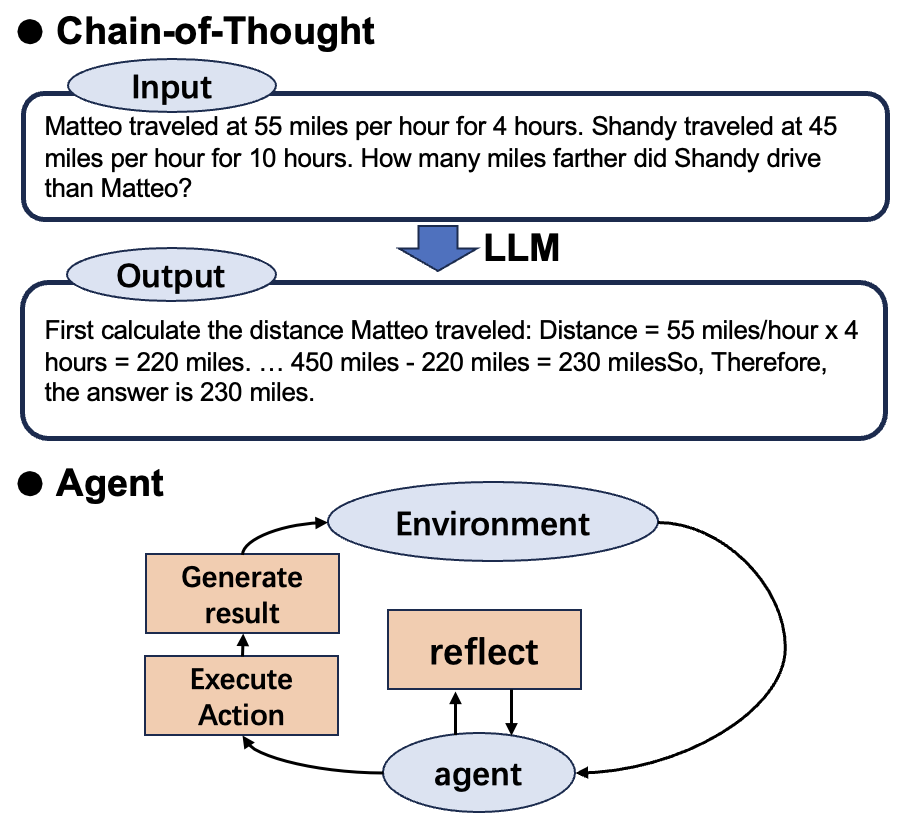}
\caption{The framework of chain-of-thought (COT) and autonomous agent. COT generally is a text paragraph, while the autonomous agent can respond multiple times to address the problem.}
\label{fig_comp}
\end{figure}

Large Language Models (LLM) have showcased remarkable performance on many tasks \cite{yao2022react,wang2023survey}, which inspires humans to consider leveraging LLM to solve challenging and complex problems.
It is worth highlighting the attention given to complex reasoning tasks.
% , consisting of commonsense reasoning, arithmetic reasoning and temporal reasoning et al.
Different from typical natural language processing (NLP) tasks, performing
complex inference requires explicitly demonstrating the analyzing process instead of simply presenting the answer, namely 
the recently proposed chain-of-thought (COT) prompting approach \cite{wei2022chain}. 
Research on COT \cite{hao2023reasoning,xie2023decomposition,diao2023active} significantly boosts the reasoning ability of LLM and achieves state-of-the-art results.

However, chain-of-thought prompting is flawed and is primarily limited by the following three constraints: 
i) \textit{hallucination issues} \cite{yao2022react,huang2023survey}, which is the main cause of COT performance degradation. 
Hallucinated reasoning is serious in COT leading to the reasoning process being seemingly plausible but lacking factual evidence. 
ii) \textit{restricted interpretability}. Although the goal of COT is to explain how an answer is yielded, it is usually presented through a text paragraph rather than a more logically organized format.
iii) \textit{uncontrollable generation}. Since COT is a one-time generated reasoning process with a large number of tokens, any mistake in the decoding will result in error perpetuation \cite{chen2022program} throughout all subsequent decoding steps.
% Prior research \cite{zhou2022least,xie2023decomposition} has proposed breaking down the source problem into a series of subproblems and solving them sequentially, which has verified the effectiveness of solving problems in a stepwise manner. Additionally, researchers present several self-evaluate \cite{xie2023decomposition} or self-confidence \cite{diao2023active} strategies at each decoding step to prevent potential error accumulation.

In this paper, our sight is to solve complex reasoning tasks based on the autonomous agent framework. 
As shown in Figure~\ref{fig_comp}, different from the typical chain-of-thought methods generating analysis process at once, agent-based approaches naturally embody the idea of step-by-step problem-solving, which addresses specific sub-problems at each step during the iterative process.
We follow the agent setup proposed by \cite{yao2022react} and design appropriate prompts to drive LLM to follow instructions.
% LLM agent 
At each step, LLM agent detects the change in the environment and conducts a response to the current state. The generated response will lead to the environment‘s change and the agent's response once again until the problem is resolved.

We further present AgentCOT to optimize the aforementioned agent setting for better adaptation to the reasoning tasks. 
Specifically, when sensing the change in the environment, AgentCOT first selects an action $a$ from a predefined set of actions and offers a specific description of the action $a_{des}$ for the current issue.
Next, AgentCOT performs the action and yields an intermediate result $R_{inter}$, while also presenting supporting evidence $E_{inter}$ for its conclusion. \{$a$, $a_{des}$,  $E_{inter}$, $R_{inter}$\} forms an atomic response state, where $a$ as well as $a_{des}$ can be viewed as a plan for the current subproblem and $E_{inter}$ can be regarded as a COT with the smallest logical unit. Such an organizational format enhances the explainability of the reasoning process.
In this way, we can carry out operations at the subproblem level, such as reflecting and redecoding, thereby achieving a comparatively controllable reasoning process. 
We propose enhanced self-consistency to enable the quality of each state, effectively preventing error perpetuation and hallucination problems. 
Additionally, we integrate the state index into the inference process to form an implicit graphical structure, which can represent a greater variety of reasoning logic.

% agentcOT是需要执行多次动作，每一次解决一个子问题，这就意味着更多的时间和资源消耗，我们提出with core 来进行推理过程的优化。

% 实验。

We evaluate the proposed approach on six common benchmarks with three types. 
From the results, AgentCOT shows competitive performance in all datasets ($\S$ \ref{sec:main}).
We further conduct experiments to compare COT framework and agent framework ($\S$ \ref{sec:1}), and carry out error analysis ($\S$ \ref{sec:2}) and case study ($\S$ \ref{sec:3}) to provide a concrete view of AgentCOT.
Finally, we conduct ablation studies to explore the model structure ($\S$ \ref{sec:4}, $\S$ \ref{sec:5}). 

In conclusion, our contributions are three-fold:
\begin{itemize}
    \item{We propose AgentCOT, a llm-based autonomous agent framework, which tackles reasoning tasks in an agent-style manner through multiple rounds of LLM generation and exhibits promising performance in different tasks.} 
    \item{To better address reasoning tasks, we organize the response of the agent at each step into a state with enriched information, containing action, action description, supporting evidence, and intermediate result. What's more, two enhancement strategies of AgentCOT are proposed to enhance performance.}
    \item{Experiments show that our method can significantly improve the state-of-the-art and is effective across various datasets and models.}
\end{itemize}

\input{doc/related}

\input{doc/approach}

\input{doc/result}

\section{Conclusion}
In this study, we present AgentCOT to alleviate the key issues faced in chain-of-thought for reasoning tasks: hallucination problem, restricted interpretability and uncontrollable generation. 
AgentCOT uses a gradual response approach to solve problems in a stepwise manner.
Each response contains action, action description, supporting evidence and intermediate result.
Experimental results on six common datasets show that AgentCOT can achieve promising performance over current competitive baselines.
The emergence of large language models sparks researchers to solve more challenging tasks.
This work employs LLM as an autonomous agent to solve reasoning tasks. We hope this work can inspire other research.

\section{Limitations}
In this paper, AgentCOT achieves state-of-the-art performance by multiple round LLM generation. 
In addition, the implementation of enhanced strategies for AgentCOT also necessitates repeated calls to the LLM, resulting in higher consumption of time and resources.
Another limitation is that AgentCOT struggles to autonomously execute the action ‘Evaluate’, requiring the development of programs to perform this action. Future research should focus on how to design prompts that enable the agent to acquire this capability.

\bibliography{arxiv}

\appendix

\section{Example Appendix}
\label{sec:appendix}

\input{doc/appendix}

\end{document}

%% file: doc/related.tex
\section{Related Work}
\subsection{LLM-based Autonomous Agent}
% \paragraph{Agent}
% [Autonomous agents have long been a prominent research focus in both academic and industry communities ~. ]
Large language models (LLM) deliver the ability to solve many challenging tasks in the real world, such as decision-making, reasoning and planning, which sparks the development of autonomous agents in human-level intelligence based on LLM \cite{wang2023survey}.
In recent times, there has been an explosive rise in applications of LLM-based intelligent agents.
For example, \citet{park2023generative} instantiate generative agents in The Sims to realize dynamical plan behavior;
\citet{li2023camel} propose a novel communicative agent framework to provide insight into cognitive processes.
% \citet{wang2023survey} have summarized the representative applications of intelligent agents in different scenarios, including social science, natural science and engineering.
% Further, \citet{wu2023autogen} categorize the agents into single-agent systems and multi-agent systems.
% AutoGen: Enabling Next-Gen LLM Applications via Multi-Agent Conversation 将agent分为
Several works focus on decision-making
agents to easily use tools, such as 
ToolLearning \cite{qin2023tool}, Reflexion \cite{shinn2023reflexion}, Toolformer \cite{schick2023toolformer}, HuggingGPT \cite{shen2023hugginggpt}, WebGPT \cite{nakano2021webgpt}. 
% In this paper, we employ LLM as an intelligent agent to guide the reasoning process. 

\subsection{Multi-Step Reasoning}
% \paragraph{Multi-Step Reasoning}
% 综述。
% Unlike traditional natural language processing (NLP) tasks, reasoning tasks require a process of 'thinking' for better resolution, namely chain-of-thought (COT) \cite{wei2022chain}, which can be seen as a multi-step reasoning process. 
% 小模型
% Previous research \cite{qu2020rnnlogic,wang2022deep,liu2023document} has already embraced such ideas for small-scale models, but always relies on a well-designed and complex model architecture.
% 大模型
The emergence of LLM enables the presentation of intermediate reasoning steps in the form of natural language \cite{zhang2022automatic}.
For current research, investigations into the COT can be segmented into three key dimensions: 
i) samples selection in in-context-learning (ICL) demonstrations \cite{mann2020language}. The perspective of choice includes diversity \cite{zhang2022automatic}, the most helpful and informative \cite{diao2023active}, relevant as well as complementary \cite{ye2022complementary}. 
% \cite{li-etal-2023-unified} propose to train a unified demonstration retriever for a wide range of tasks.
% Additionally, the organization of selected examples also influences the performance of ICL \cite{rubin2021learning}, such as the order.
% aThis is also the core of ICL research.
ii) the refinement of the COT. 
\citet{fu2022complexity} show
that superior COT with higher reasoning complexity.
\citet{zhou2022least} propose a pipeline method to first provide a plan to break down the source problem into several subproblems and then solve them sequentially.
Subsequent works \cite{xie2023decomposition,hao2023reasoning,besta2023graph} discard the planning phase to prevent the impact of planning errors on problem resolution. They always present self-evaluation strategies to improve the correctness of each step, such as stochastic beam search \cite{xie2023decomposition}, self-confidence \cite{diao2023active} and self-consistency \cite{wang2022selfconsistency}.
iii) the reflection and verification after generating COT, aiming at recognizing issues in produced COT based on LLM \cite{wang2022selfconsistency,xie2023decomposition,kim2023language} or extra tools \cite{gou2023critic}.
% , and regenerating a new one without the discovered problems.
% ==== 
% In the work, we present a novel perspective to generate agent-style  COT based on the LLM agent.

%% file: doc/approach.tex
\begin{figure*}[ht]
\centering
\includegraphics[scale=0.22]{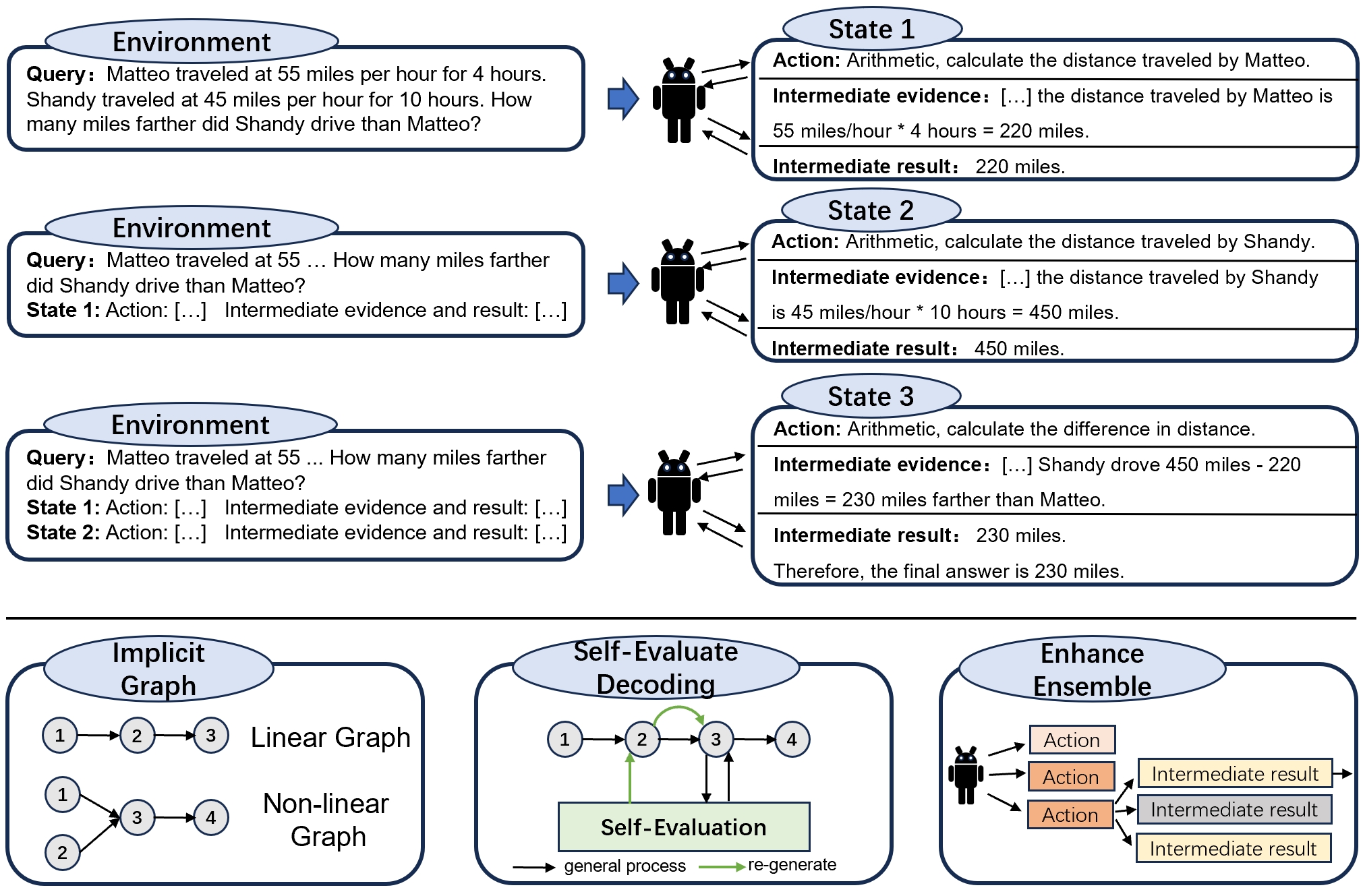}
\caption{The overview of our method AgentCOT. An instance of AgentCOT's execution process is visualized at the top of the figure. At each step, LLM agent senses the change in environment and generates action, action description, intermediate evidence, and intermediate result sequentially. These pieces of information with efficient organizations respond to the environment and result in the environment changing once again. We also provide some details for the implicit state graph, self-evaluate decoding and enhanced ensemble strategy at the bottom.}
\label{fig_model}
\end{figure*}

\section{Approach}
We propose AgentCOT to treat LLM as an autonomous agent to perform textualized agent-style reasoning, 
% guide the reasoning process and generate chains of thought gradually, 
which is illustrated in Figure~\ref{fig_model}.
The model consists of two important components: 1) Agent Solving (\S 3.1), which is the foundational framework of AgentCOT for addressing reasoning problems. 2) AgentCOT Enhancement (\S 3.2), which involves multiple techniques to improve the effectiveness of our method.
The simplified prompt for AgentCOT is shown below and the full prompt scheme is presented in Appendix~\ref{sec:prompt_design}.
\includegraphics[width=0.48\textwidth]{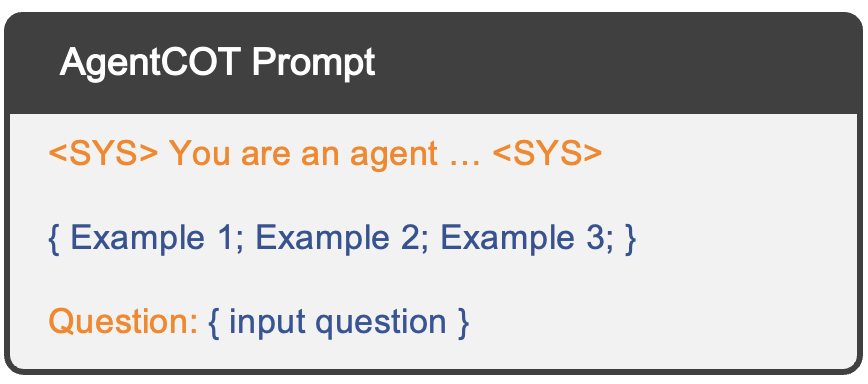}

\subsection{AgentCOT Solving}
\paragraph{LLM as Agent}
Inspired by previous work that integrates reasoning and acting advances \cite{yao2022react}, we develop the agent setup for reasoning tasks, which interacts with the environment $E$ after perceiving the change in $E$ and then taking action $a$ responding to $E$. 
% 智能体的一般执行过程。
Specifically, the initial environment $E$ only includes the original query $Q^0$. At step $i$, the agent senses the change in $E$ with state $Q^i$, which drives itself to execute the action $a^i$ in action set $\mathcal{A}$ and conducts the result $r^i$. The $Q^i$ and $r^i$ are concatenated to form  $Q^{i+1}$, which leads to the environment‘s change and the agent's response once again until the problem is resolved.
The prompt presented below is designed to enable the LLM to act as an agent.
\includegraphics[width=0.48\textwidth]{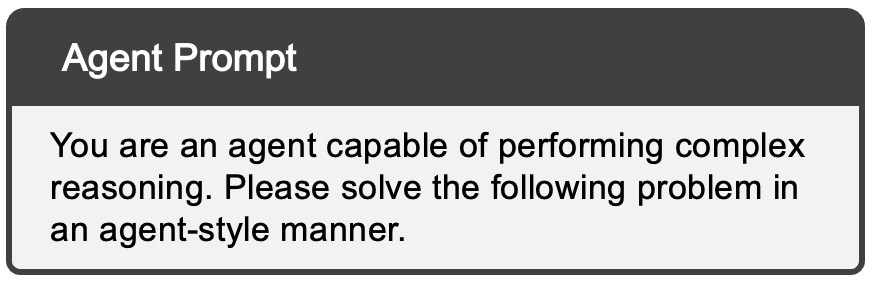}

\paragraph{Action Space and Action Selection}
% 有限动作集 
The action space in our method comprises a finite set of actions $\mathcal{A}$ related to question-answering reasoning tasks.
% 从艾伦中选， 
We include all actions brought forward by \cite{allenaction} within our action set, containing 13 operator types: Select, Filter, Project, Aggregate, Group, Superlative, Comparative, Union, Intersection, Discard, Sort, Boolean and Arithmetic.
% 额外定义一个重要的描述。
We further define the action \textit{Describe}, which explains nouns, states, or actions; the action \textit{Evaluate} to assess the quality of generated information. 
% These two additional actions will be described in detail in $\S$ 3.2. 
% agent将从上述动作集中选择动作，
When detecting changes in the environment, the agent will select an appropriate action $a$ from $\mathcal{A}$. Presented below is the prompt for the action set.
\includegraphics[width=0.48\textwidth]{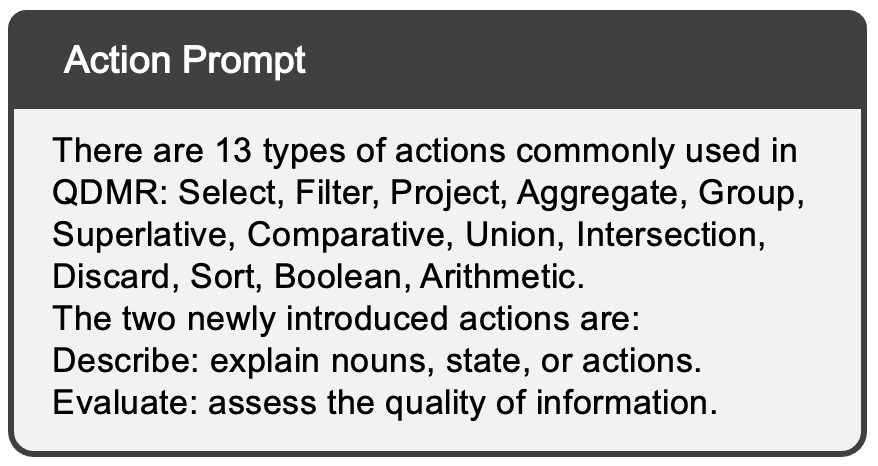}

% AgentCOT除了给出选择动作之外，还会让给出执行动作的具体说明，来更明确告诉执行体要做什么。
When solving reasoning problems, AgentCOT not only presents the option for actions but also delivers a detailed description $a_{des}$ of the selected action $a$, ensuring clear instruction during execution.
% 考虑到我们的动作集可能是不全面的，而且一些动作可能不需要明确定义，如不常见的动作，我们也允许模型不选择动作，而是直接给出执行动作的说明。
What's more, considering our action set $\mathcal{A}$ may be incomplete and some actions may not necessitate definitions, we also allow the agent not to make an action selection and, instead, to only supply a detailed description of what needs to be executed.

\paragraph{Action Executing}
% AgentCOT允许动作执行体是LLM或者其他的外挂工具，如代码编辑器，搜索引擎等。 
AgentCOT allows for the executor of the selected action to be LLM itself or other external tools, such as search engine or calculator.
% 当agent与环境交互给出动作后，执行体会执行动作指令，并给出相应的答案。
After the agent interacts with the environment $E$ and provides an action $a$,  the executor will execute the action $a$ and produce the corresponding results.
% 当使用LLM大模型作为执行体，我们要求大模型给出中间证据和中间答案。
When employing LLM as the executor, we require the model must provide intermediate evidence $E_{inter}$ and intermediate result $R_{inter}$.
% 中间证据指的是由动作给出中间结果的分析过程，其可以被认为一个解决执行动作解决子问题的最小思维链。
$E_{inter}$ refers to the analysis process in which action and action description generate intermediate results, which can be regarded as a minimal chain of thought to solve the current subproblem.
% 中间答案指的是执行动作后得到的结果。
$R_{inter}$ means the result obtained following action instruction.
Other tools as executors only need to provide the intermediate result $R_{inter}$,
% 其他工具作为执行体只需要提供最终答案，
% （或者再次经由LLM进行答案后处理，如搜索外挂工具可以无法）

\paragraph{Enriched State and Implicit State Graph}
% 根据上面的描述，当agent与外界进行交互之后，相当于会产生一个含有丰富信息的状态，包含：动作，动作的解释，中间证据，中间结果。
As described above, at each step $i$, after sensing the change in the environment, AgentCOT will generate an information-rich state $S^i$, encompassing action, action description, intermediate evidence and intermediate answer:
\begin{align}
  S^i = \{a^i, a_{des}^i, E_{inter}^{i}, R_{inter}^{i}\}
\end{align}  
% 我们通过实验证明了这种含有丰富信息的节点会更加有效和缓解幻觉现象。
Experimental results have demonstrated that a state with extensive information can support superior performance.

% 虽然每一个状态只能逐个的生成，但是这并不代表状态之间的关系是线性的，如XX。  
Further, although states are generated one by one, it does not imply that the interrelationships between states are \textit{chained}. For example, the first state and second state are independent, while the third node relies on both the first and the second simultaneously, as shown in the first figure at the bottom of Figure~\ref{fig_model}. 
% 为了表示这种复杂的推理模式，我们将状态的下标也融入到状态的信息中，从而加强各个状态之间的联系，也就是说在agentcot解决问题时包含一个隐式的图结构。
To depict this complex reasoning pattern, we integrate the state index into the state itself, thereby strengthening the connections between states. 
Specifically, the state indexes mainly exist in $a_{des}^{i}$ and $E_{inter}^{i}$. When $E_{inter}^{i}$ needs to contain information in $E_{inter}^{j} (j<i)$, the corresponding information in $E_{inter}^{i}$ will be written as '\# j', or an additional '(\# j)' will be added after the corresponding information.
Therefore, essentially, AgentCOT encompasses an implicit graphical structure when solving problems.

\paragraph{Iterative Process}
% 迭代过程
After producing an information-rich state $S^i$ at step $i$, the question in $E$ will be updated as follows:
\begin{align}
  Q^{i+1} = Q^i + S^i
\end{align}
% For AgentCOT, we reorganize core information instead of employing all in $S^i$.
% Intuitively, subsequent steps of problem-solving only require previous actions $a$, action descriptions $a_{des}$ and intermediate results $R_{inter}$, without the need to consider how the $R_{inter}$ was obtained, namely:
% \begin{align}
%   Q^{i+1} = Q^i + \{a^i, a_{des}^i, R_{inter}^{i}\}
% \end{align}
which will result in the agent's response again. AgentCOT iteratively executes until it generates the final result. The final result is typically the outcome of the last action taken, presented as 'Therefore, the final answer is ...'.
% 终止条件。   
% The prompt for executing the above agent-style reasoning is shown below.
The detailed prompt for supporting the LLM in performing agent-style reasoning is shown below.
\includegraphics[width=0.48\textwidth]{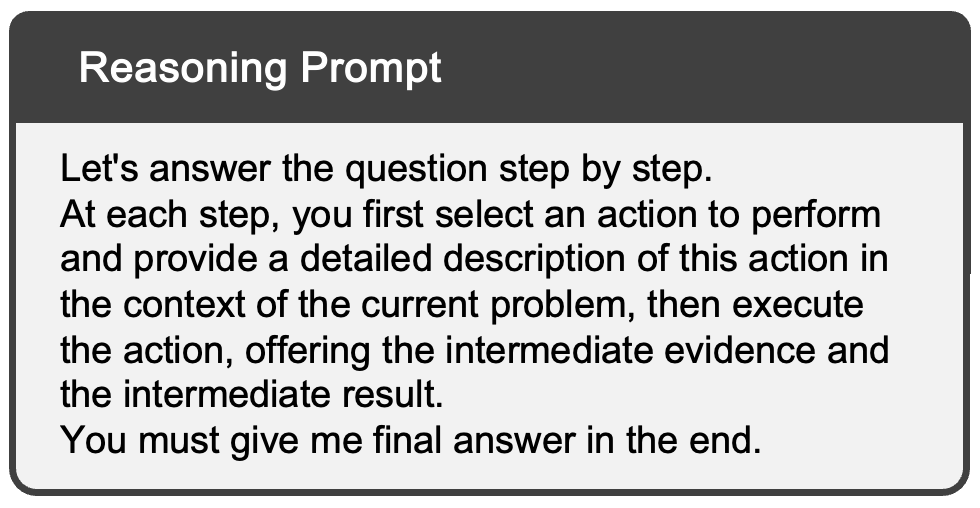}

\subsection{AgentCOT Enhancement}
As described above, AgentCOT demonstrates an explicit multi-step reasoning process.
Inspired by \cite{xie2023decomposition}, our proposed enhanced strategy is based on each step generated. As shown in the second figure at the bottom of Figure~\ref{fig_model}, AgentCOT can evaluate the quality of each generated state and decide whether to continue reasoning or go back to regenerate. Evaluation and reflection essentially provide a solution to the non-reversible issue in decoding strategy for current LLM.

AgentCOT ensures the state quality at two levels. The first is the subproblem level.
We employ a divergent thinking strategy to allow multiple different reasoning paths. Specifically, AgentCOT generates multiple responses each time. Then, we perform ensemble learning by considering both actions and intermediate results to select the optimal response, as presented in the third figure at the bottom of Figure~\ref{fig_model}.
The second is the global problem level. AgentCOT is easy to convert into the COT paradigm with enriched information. 
At each decoding step, we encourage AgentCOT to generate the remaining complete inference process, which means that AgentCOT will generate a final result at this point to help evaluate generated states. 
As a result, in every response, AgentCOT considers two levels simultaneously, namely containing actions, intermediate results and suggestive final results, to provide the best state. 
% The implementation details are presented in the 

% \subsection{Compare to previous work}
% % 介绍一些我们框架的特点：
% % 1）含有更加丰富信息的状态：  
% % 1.动作的解释，2.引入中间证据而不是直接给出答案。3. 状态允许非线性的，而是以隐式的图的形式存储。
% % 2）

% But different from the previous method \cite{yao2022react, shinn2023reflexion}, we further modified the framework to optimize its performance in question-answering reasoning tasks.
% % 我们对其进行修改使得其可以更好的适用于问答推理任务。
% % 1）一个更加智能的框架，样例选择，问题解决和反思。
% % 2）对问题解决进行了更好的优化。引入样例选择和反思。

%% file: doc/result.tex
\section{Experimental Setups}
% 数据集
% \paragraph{Datasets and Evaluation Metrics.}
\subsection{Datasets and Evaluation Metrics}
We conduct experiments on six common benchmarks, which can be classified into three categories: (1) arithmetic reasoning,  containing \textbf{GSM8K} \cite{cobbe2021training} and \textbf{AQuA} \cite{ling2017program}. (2) commonsense reasoning, including \textbf{CommonsenseQA} \cite{geva2021did} and \textbf{Date} \cite{wei2022chain}. (3) multi-hop question answering based on fact, consisting of \textbf{Bamboogle} \cite{press2022measuring} and \textbf{Compositional Celebrities} \cite{press2022measuring}.
Table~\ref{tab:dataset} shows their detailed statistics.
% 指标
Following the previous work \cite{wei2022chain,zhang2022automatic}, we report accuracy as evaluation metrics for all datasets.
\subsection{Implementations.}
% 大模型：
For the large language model, we mainly leverage two versions of GPT \cite{brown2020language}, \textit{text-davinci-002} and \textit{gpt-3.5-turbo}, to conduct experiments.
% and we also introduce other LLM to perform an ablation study, such as ERNIE Bot \url{https://yiyan.baidu.com/}, Tongyi Qianwen \url{https://qianwen.aliyun.com/} and Llama \cite{touvron2023llama}.
% open source model  . 
% prompt：数量
In our implementation, we select several examples from the training dataset, if available, to form demonstrations \cite{brown2020language} for in-context learning. The number of examples are following previous works \cite{wei2022chain,diao2023active}. 
% We present some constructed demonstrations of our method in Appendix X.
% 超参
For the hyper-parameters in the inference stage, the temperature is chosen from \{0.8, 0.9, 1.0, 1.1, 1.2\}, and the top-p value is selected in \{0.8, 0.9, 1.0\}.
The maximum number of calls for LLM when performing enhanced strategy for AgentCOT is set from \{3, 4, 5\}.
\subsection{Baselines.}
We compare AgentCOT with several baselines as follows: 
\textbf{COT} \cite{wei2022chain}, the first paper proposing chain-of-thought. 
\textbf{COT-SC} \cite{wang2022selfconsistency} generates COT based on self-consistency decoding strategy. 
\textbf{Auto-COT} \cite{zhang2022automatic} shows an automatic COT prompting approach that considers diversity in the demonstrations. 
\textbf{Complex-COT} \cite{fu2022complexity} is inclined to choose the COT that includes a higher count of reasoning steps.
\textbf{Random-COT} \cite{diao2023active} randomly selects examples from the training set to form demonstrations.
\textbf{PAL} \cite{xie2023decomposition} introduces self-evaluation guided beam search to enhance the COT.
\setlength{\tabcolsep}{2.2pt}
\begin{table}[t]
\centering
\small
\begin{tabularx}{\linewidth}{Xcccccc}
\toprule
\textbf{} & \textbf{GSM8K}  & \textbf{AQUA} & \textbf{CSQA} & \textbf{Date} & \textbf{Bamboogle} & \textbf{CC}\\ \midrule 
\textbf{Train} & 7,473 & 254 & 12,247 & - & - & -  \\
\textbf{Dev} & - & 254 & 1,221 & - & - & -  \\
\textbf{Test} & 1,319 & 404 & 1,140 & 369 & 125 & 8,693   \\
\bottomrule
\end{tabularx}
\caption{Statistics of datasets.}
\label{tab:dataset}
\end{table}
\setlength{\tabcolsep}{2.3pt}
\begin{table*}[t]
\centering
\begin{tabularx}{\linewidth}{Xlcccccc}
\toprule
\textbf{Model} & \textbf{Method} & \textbf{GSM8K} & \textbf{AQuA} & \textbf{CSQA} & \textbf{Date} & \textbf{Bamboogle} & \textbf{CC}\\ \midrule
\multirow{7}{*}{{text-davinci-02}} 
& COT \cite{wei2022chain} & 46.9* & 35.8* & 73.5* & 52.1* & 32.8 & 44.3 \\
& COT-SC \cite{wang2022selfconsistency} & - & - & - & - & 36.0 & 46.2 \\
& Auto-COT \cite{zhang2022automatic} & 47.9* & 36.5* & 74.4* & - & - & - \\
& Complex-COT \cite{fu2022complexity} & 55.4* & 37.8 & 73.7 & 59.0 & 48.8 & 47.7 \\
& Random-COT \cite{diao2023active} & 63.9 & 44.1* & 74.5* & 62.2 & 50.4 & 47.2 \\
% & Active-COT \citeyearpar{diao2023active} & 71.1* & 48.4* & 76.6* &  &  &  \\
& PAL \cite{xie2023decomposition} & 58.1 & 35.2 & 74.9 & 59.6 & 51.2 & 54.7  \\
\cline{2-8} 
& Agent-COT (Ours) & 67.1 & 38.6 & 78.4 & 64.1 & 52.0 & 57.6 \\
\hline
\multirow{6}{*}{{gpt-3.5-turbo}} 
& COT \cite{wei2022chain} & 73.8 & 57.0 & 71.3 & 58.2 & 56.8 & 55.2 \\
& COT-SC \cite{wang2022selfconsistency} & 75.4 & 58.6 & 72.9 & 59.8 & 58.3 & 57.1 \\
& Complex-COT \cite{fu2022complexity} & 71.9 & 57.8 & 72.9 & 58.8 & 55.2 & 57.6 \\
& Random-COT \cite{diao2023active} & 75.3 & 55.5 & 73.7 & 61.2 & 56.8 & 56.6 \\
& PAL \cite{xie2023decomposition} & 72.7 & 55.5 & 64.7 & 62.6 & 56.8 & 55.3 \\
% \cmidrule(lr){2-8} 
\cline{2-8} 
& Agent-COT (Ours) & 79.9 & 59.8 & 79.5 & 64.4 & 58.4 & 58.5 \\
\bottomrule
\end{tabularx}
\caption{Overall results of our approach compared to previous works on different datasets with three task types.  * means
the result is from the original paper.}
\label{tab:main}
\end{table*}
\section{Experimental Results}
\label{sec:main}
We present the main experimental results of our method compared to strong baselines in Table~\ref{tab:main}, which contain six datasets with three types and two versions of GPT model.
From the results, we can find that our method AgentCOT achieves the best performance over most datasets and different versions of GPT. AgentCOT beats COT \cite{wei2022chain} by increasing 12.06\% and 4.70\% accuracy on average in \textit{text-davinci-002} and \textit{gpt-3.5-turbo} respectively, which has verified the superiority of agent framework over traditional COT.
% 不同版本的模型
Due to the better model capability on upgraded version \textit{gpt-3.5-turbo} than \textit{text-davinci-002}, our method and baselines obtain higher results in \textit{gpt-3.5-turbo}, particularly on arithmetic reasoning datasets GSM8K and AQuA. For our method, AgentCOT demonstrates nearly comparable performance on two versions of GPT model on CSQA, Date and Bamboogle, indicating that our carefully designed agent framework effectively activates the problem-solving capabilities of the model, thereby bridging the gap in original ability.
% 不同类型的数据集。
By comparing AgentCOT with baselines in different types of datasets, we can see there are significant discrepancies in the improvements AgentCOT gained. 
Taking the results on \textit{text-davinci-002} as an example, overall, AgentCOT shows the highest increase on the multi-hop question answering dataset (+16.7\% on average), followed by arithmetic reasoning (+11.5\% on average), and finally commonsense reasoning (+7.9\% on average).
A reasonable interpretation is that there are clear boundaries in step-by-step execution for multi-hop question answering and arithmetic reasoning, which can be completed based on AgentCOT's ability of problem decomposition.
The results on two commonsense reasoning datasets with different natures also exhibit considerable differences. Through further analysis, CSQA is a dataset for reasoning about everyday life scenarios, while Date is about date calculations, which is more suitable for the step-by-step problem-solving approach of the AgentCOT framework.
\section{Discussion}
In this section, we conduct a series of detailed studies to explore AgentCOT's ability.
\subsection{COT framework or Agent framework?}
\label{sec:1}
\begin{figure}
  \subfigure{
    \includegraphics[width=0.45\textwidth]{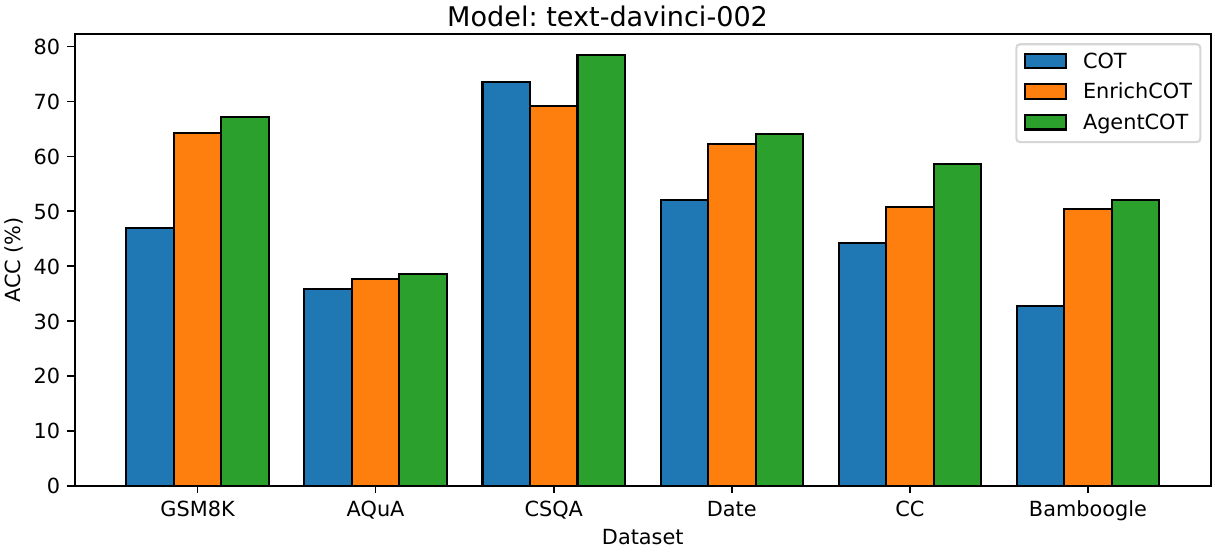}
  }
  \subfigure{
    \includegraphics[width=0.45\textwidth]{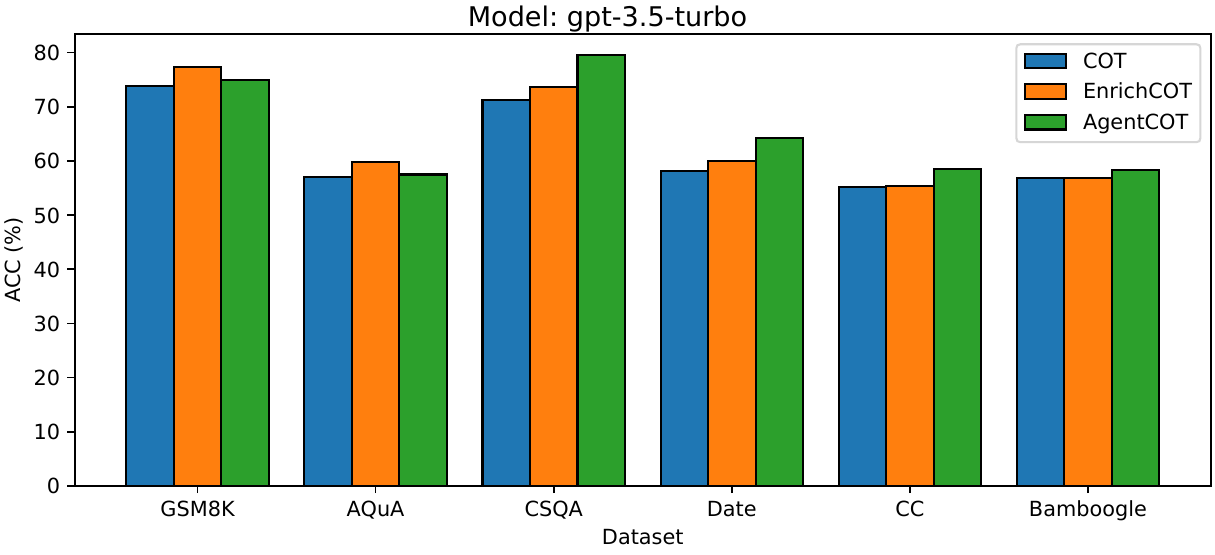}
  }
  \caption{Performance comparison between COT paradigm and agent paradigm. 'COT' denotes the chain-of-thought proposed by \cite{wei2022chain}. 'EnrichCOT' is to consider the reasoning process of AgentCOT as a one-time generation of COT.}
\label{fig_enrich}
\end{figure}

% 我们的方法可以退化成含有丰富信息的思维链的形式，我们称之为EnrichCOT。
% 我们在6个数据集上比较普通的COT模式和agent模式的性能。
Our method AgentCOT can be degraded into the general COT paradigm with enriched information, which we call EnrichCOT. Figure~\ref{fig_enrich} shows the comparison of the performance of the COT framework and the agent framework on six datasets in \textit{text-davinci-002} and \textit{gpt-3.5-turbo} respectively.  
From the results, we can find that AgentCOT significantly outperforms EnrichCOT in most datasets. 
% agentCOT的效果是明显优于enrichCOT的，可能的原因是基于代理框架的agentcot具有具有更加可控的推理过程，在每一步生成新的状态时都会有策略确保生成的质量。
A reasonable explanation is that AgentCOT, grounded in an agent framework, provides a more controlled inference process, implementing effective strategies to ensure the quality of generated states at each step.
We also notice that EnrichCOT achieves higher accuracy on GSM8K and AQuA in \textit{gpt-3.5-turbo}, which indicates explicit problem deposition can disrupt the process of thinking for the arithmetic reasoning task.
% 与COT相比，EnrichCOT的性能更好，这说明丰富的信息如动作等对推理过程是有帮助的。
Compared to COT \cite{wei2022chain}, EnrichCOT demonstrates superior performance, suggesting that enriched information, such as actions, intermediate evidence, and intermediate result, proves beneficial to help reasoning.

\begin{figure}[ht]
\centering
\includegraphics[scale=0.13]{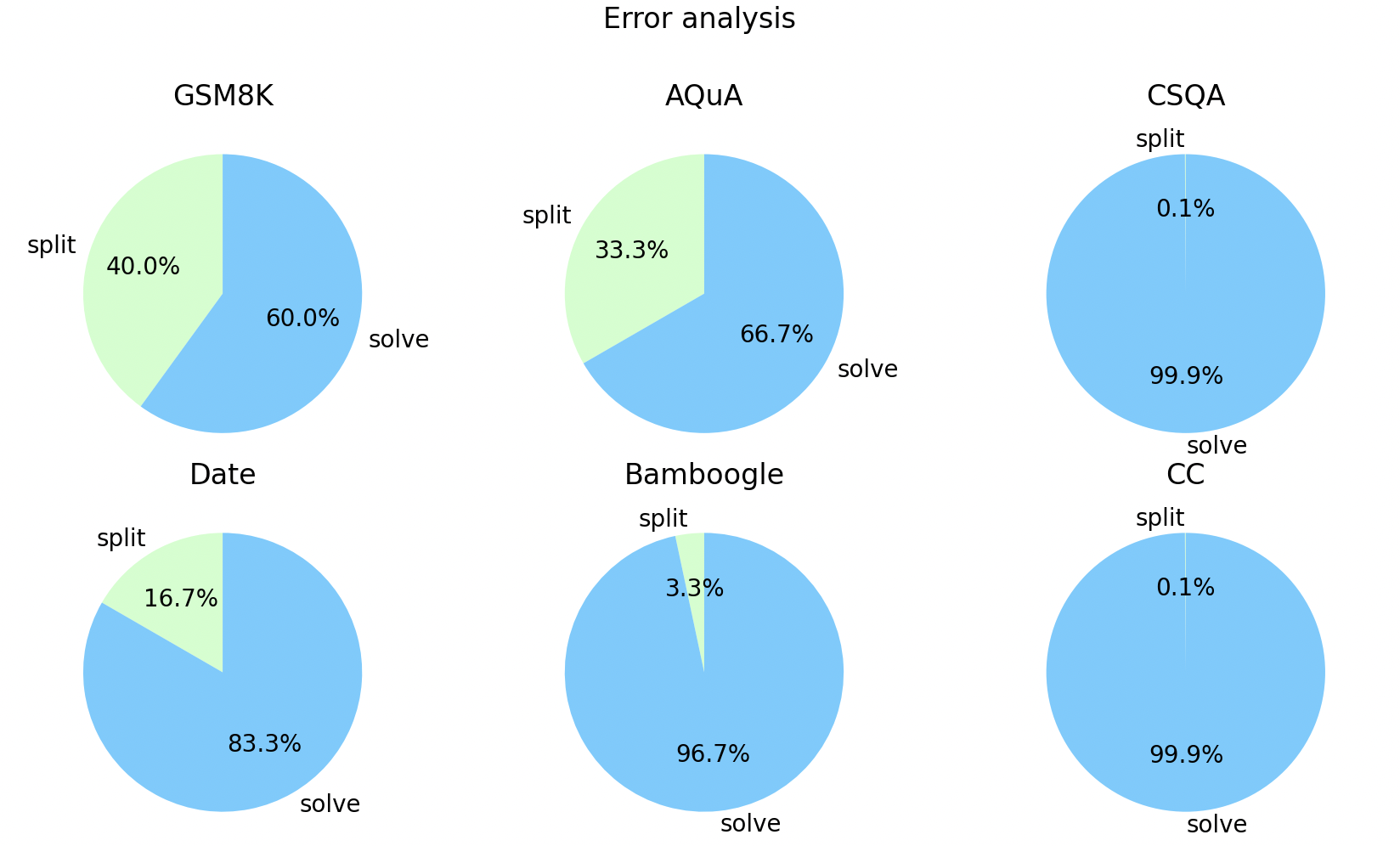}
\caption{Error Analysis for exploring the ability of AgentCOT. The percentages of examples in which problem decomposition errors ('split') and subproblem solution errors ('solve') occur during the inference process are given in six datasets.
}
\label{fig_error}
\end{figure}

\subsection{Error Analysis}
\label{sec:2}
We conduct error analysis to explore the lack of capability of our method on six datasets based on the model gpt-3.5-turto. 
% 我们将产生错误结果的原因分为模型分解问题能力缺失（即，动作和动作说明的发生错误）和模型解决子问题能力缺失（即，中间证据和最终答案错误）两种。
Specifically, we classify the factors leading to the erroneous reasoning process into two groups: the model's lack of problem decomposition capability (i.e., errors in actions and action descriptions) and the model's lack of subproblem-solving capability (i.e., errors in intermediate evidence and answers).
% 我们基于gpt3.5在6个数据集上进行统计，结果在图XX中。
The results are presented in Figure~\ref{fig_error}.

% 从图中我们可以得出结论：
From the percentage of samples presented in the figure, we can conclude that: 
% 1）总体上agentCOT的问题分解能力强于解决子问题的能力。通过进一步的分析，我们发现解决子问题是包含计算错误，检索知识错误等，可以借助于外挂技术进一步进行优化。
1) overall, AgentCOT demonstrates superior performance in problem decomposition compared to its ability to solve subproblems. Further investigation reveals that errors in solving subproblems mainly include computation errors and knowledge retrieval inaccuracies, which can be optimized by introducing external tools.
% 2）AgentCOT的能力在不同类型数据集上的表现是有差异的，对于常识推理（XX，XX）和多跳推理任务，问题分解能力已经足够。由于计算推理问题相对复杂，agentcot的分解问题能力表现一般，但依然强于解决子问题的能力。
2) AgentCOT's capabilities exhibit variability on different dataset types. For commonsense reasoning tasks (CSQA and Date) and multi-hop question-answer tasks (Bamboogle and CC), problem decomposition errors almost never happen. However, due to arithmetic reasoning problems being more complex, the performance of AgentCOT's problem decomposition is moderate in GSM8K and AQuA but still superior to subproblem-solving.

\setlength{\tabcolsep}{3.5pt}
\begin{table}[t]
\small
\centering
\begin{tabularx}{\linewidth}{Xccccc}
\toprule
\textbf{Setting} & \textbf{GSM8K}  & \textbf{AQuA} & \textbf{CSQA} & \textbf{Date}  & \textbf{CC}\\ \midrule 
Full Model & 79.9 & 59.8 & 79.5 & 64.4 & 58.5  \\
\hline
\ \  w/o \  Action & 78.5 & 52.5 & 74.1 & 60.7 & 51.6 \\
\ \  w/o \  ActionD & 78.4 & 55.0 & 77.0 & 61.7 & 40.7 \\
\ \  w/o \  IEvidence & 71.2 & 50.7 & 59.0 & 60.4 & 56.2 \\
\bottomrule
\end{tabularx}
\caption{Ablation study on AgentCOT framework. 'ActionD' stands for action description and 'IEvidence' refers to intermediate evidence. We conduct experiments on \textit{gpt-3.5-turbo}.}
\label{tab:ablation}
\end{table}

\subsection{Case Study}
\label{sec:3}
\begin{figure*}[ht]
\centering
\includegraphics[scale=0.21]{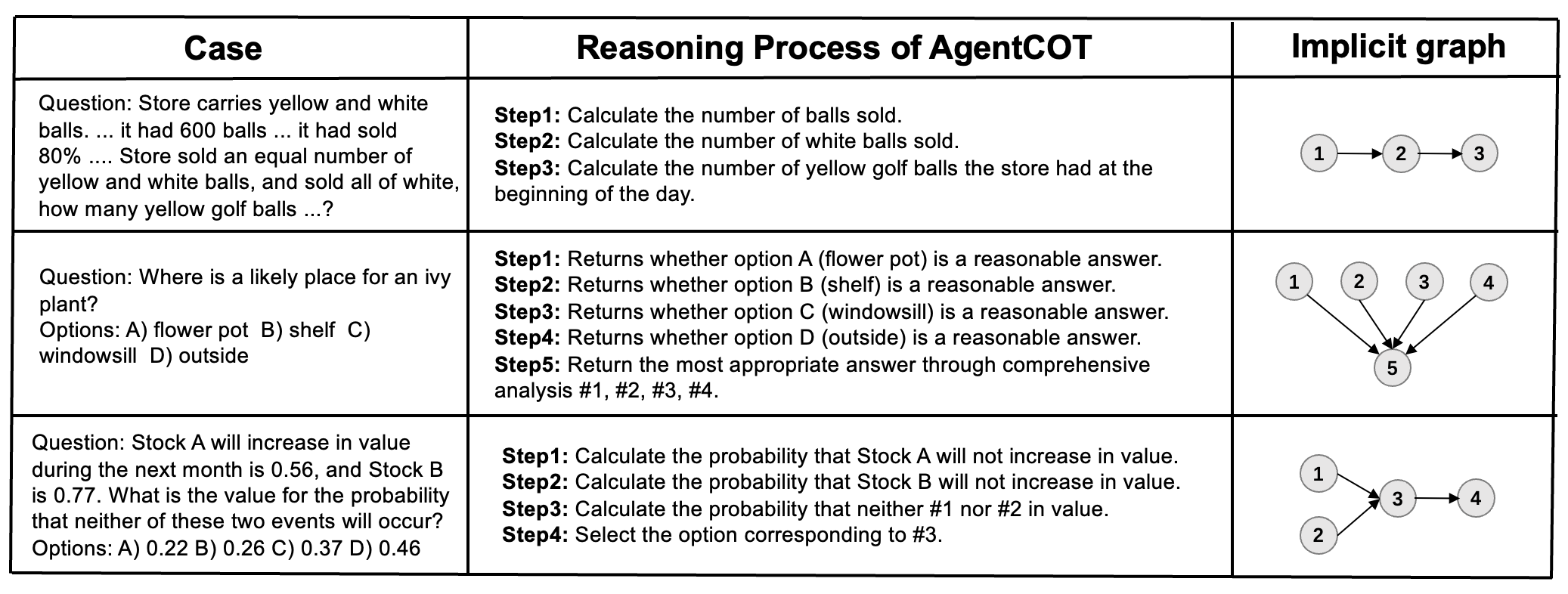}
\caption{Case study. We only provide action descriptions for clarity in the reasoning process, omitting other information. The node $i$ in the implicit graph corresponds to the Step $i$ of AgentCOT in the reasoning process and '\#$i$' indicates the use of information is from Step $i$.}
\label{fig_case}
\end{figure*}

We list three examples for the case study to provide a concrete view of different implicit graph structures in Figure~\ref{fig_case}. 
The implicit graph depicted in Case 1 is a fundamental linear structure, whereas the graphs in Case 2 and Case 3 exhibit distinct ways of node connections.
Specifically, the first case is selected from AQuA dataset. AgentCOT relies on the calculation of the previous step to obtain the outcome at each step.
The second case is chosen from CSQA dataset. For the given question, AgentCOT independently analyzes each option and then combines the analyses to yield a final answer.
The third case is selected from AQuA dataset. 
AgentCOT first calculates the probabilities of A and B stocks not increasing respectively, and then computes the probability of both of them happening. Finally, based on the calculations, AgentCOT selects the correct option.
Diverse graph structures reflect the multitude of thoughts 
 adopted by AgentCOT in problem-solving.
Such implicit graphs offer a twofold advantage. Firstly, it enhances the interpretability of the reasoning process, resulting in more easily comprehensible inference pathways. 
Secondly, it also strengthens the model itself by enabling a more explicit organization and use of information during the reasoning process.
% 第一个例子中的图是基本的线性图结构，case 2 和 case 3的图则是有不同的节点连接方式，体现了解决推理问题时，agentcot解决问题更加多样的思维方式。
% agentCOT这种隐式的图结构，一方面可以增强推理过程的可解释性，能够产生更容易理解的推理过程。另一方面，也是对其本事的一种增强，隐式的图结构使得模型本身在进行推理时更加明确的组织和使用信息。

\subsection{Ablation Study on AgentCOT Structure}
\label{sec:4}
We conduct an ablation study to explore the effect of action, action description and intermediate evidence on the performance of AgentCOT.
We carry out experiments in the version of \textit{gpt-3.5-turbo} on five different benchmarks and we report the results on Table~\ref{tab:ablation}.
From the table, AgentCOT without action results in a 4.95\%  reduction in results on average and AgentCOT without action description leads to the performance degrade about 5.89\%. 
Results indicate that actions and descriptions of those actions are both essential during the process of inference. 
Model performance significantly degrades when 
AgentCOT lacks the action description compared to the lack of action, since the action set is the same between different questions, while the action description is problem-specific and can guide problem-solving.
AgentCOT without intermediate evidence is similar to the approach proposed by \cite{xie2023decomposition}, which results in a decrease in accuracy by 8.93\%.
In fact, the intermediate evidence can be viewed as the reasoning process of the sub-problem. Such a chain of thought can help gain correct results.

\subsection{AgentCOT with Enhanced Self-Consistency}
\label{sec:5}
\begin{figure}
\centering
\subfigure[Results on CC.]{\includegraphics[width=3.7cm]{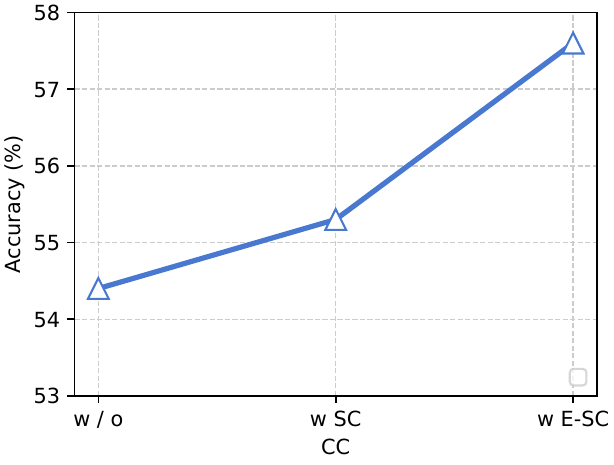}} 
\subfigure[Results on AQuA.]{\includegraphics[width=3.7cm]{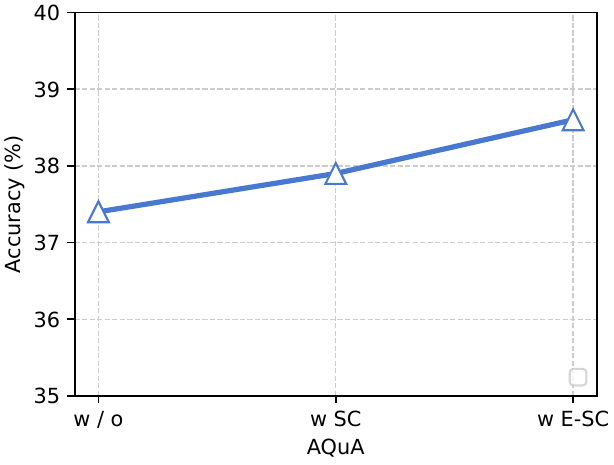}}
\caption{The results of self-consistency approaches. 'w / o' means AgentCOT without self-consistency strategies. 'w SC' and 'w E-SC' indicate AgentCOT with self-consistency strategies proposed by \cite{wang2022selfconsistency} and us respectively.} 
\label{fig:sc}
\end{figure}

In this sub-section, we evaluate the effectiveness of our proposed enhanced self-consistency.
We present the results on CC and AQuA in Figure~\ref{fig:sc}.
% Due to the large language model adopting the greedy coding strategy, 
The LLM always generates different outputs each time due to the influence of the decoding strategy.
The method proposed by \cite{wang2022selfconsistency} chooses the final answer with high confidence based on an ensemble strategy, which can provide an increase in accuracy.
AgentCOT with enhanced self-consistency strategy considers a fine-grained level to guarantee the quality of each generated step by ensembling the actions and intermediate results. 
From the results, AgentCOT with the enhanced self-consistency strategy further improves model performance by a significant margin.

% \setlength{\tabcolsep}{3.2pt}
% \begin{table}[t]
% \centering
% \begin{tabularx}{\linewidth}{Xccc}
% \toprule
% \textbf{Model} & \textbf{Statistics} & \textbf{CSQA}  & \textbf{CC}\\ \midrule 
% \multirow{2}{*}{AgentCOT} & Accuracy & 79.5 & 58.5 \\
% & Token Num & 235 & 45 \\
% \hline
% \multirow{2}{*}{AgentCOT w Core Information} & Accuracy & 79.4 & 57.8 \\
% & Token Num & 95 & 22 \\
% \bottomrule
% \end{tabularx}
% \caption{Performance of AgentCOT with different iteration information. 'Core Information' consists of action, action description and intermediate result. }
% \label{tab:core_info}
% \end{table}

% 因此，我们建议使用core information 而不是之前全部的信息，which 可以更快地，消耗资源更少的得到可比较的结果。

%% file: doc/appendix.tex
\subsection{Prompt Design}
\label{sec:prompt_design}
\begin{figure}[ht]
\centering
\includegraphics[width=0.48\textwidth]{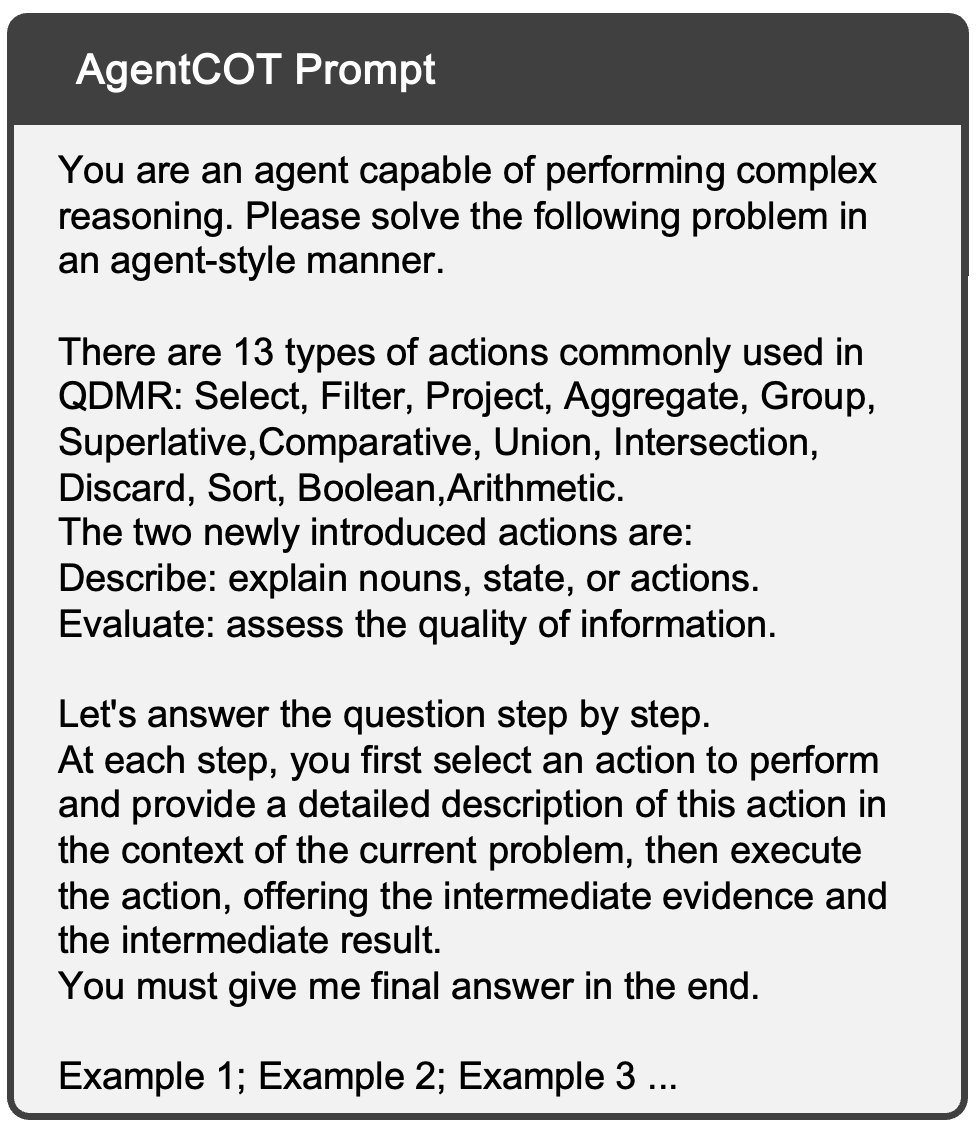}
\caption{The complete prompt for AgentCOT.}
\label{fig_full_prompt}
\end{figure}

\begin{figure*}[ht]
\centering
\includegraphics[scale=0.47]{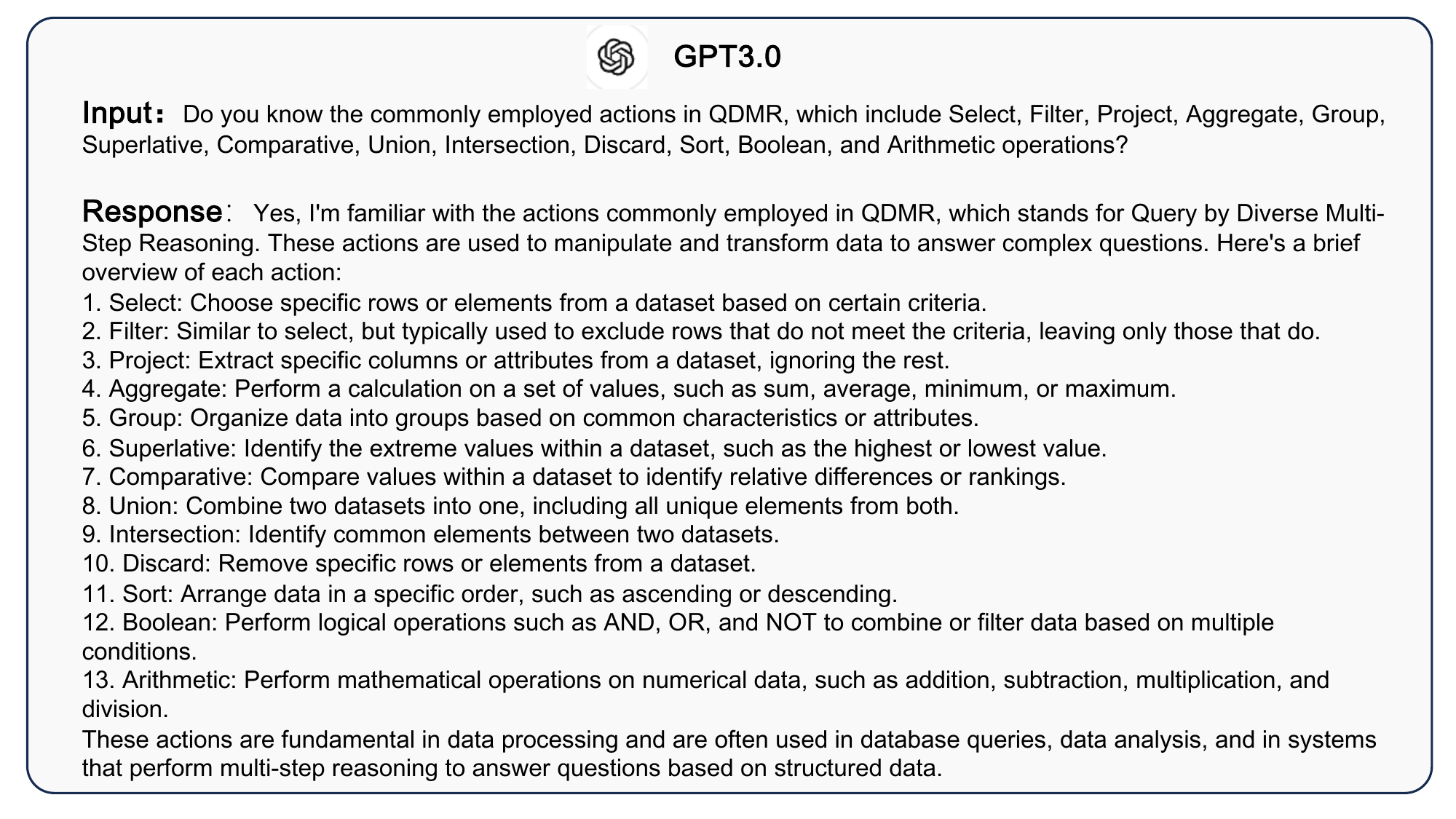}
\caption{The demonstration that the GPT-3 model includes the knowledge of actions within QDMR.}
\label{fig_gpt3_action}
\end{figure*}

\begin{figure*}[ht]
\centering
\includegraphics[scale=0.47]{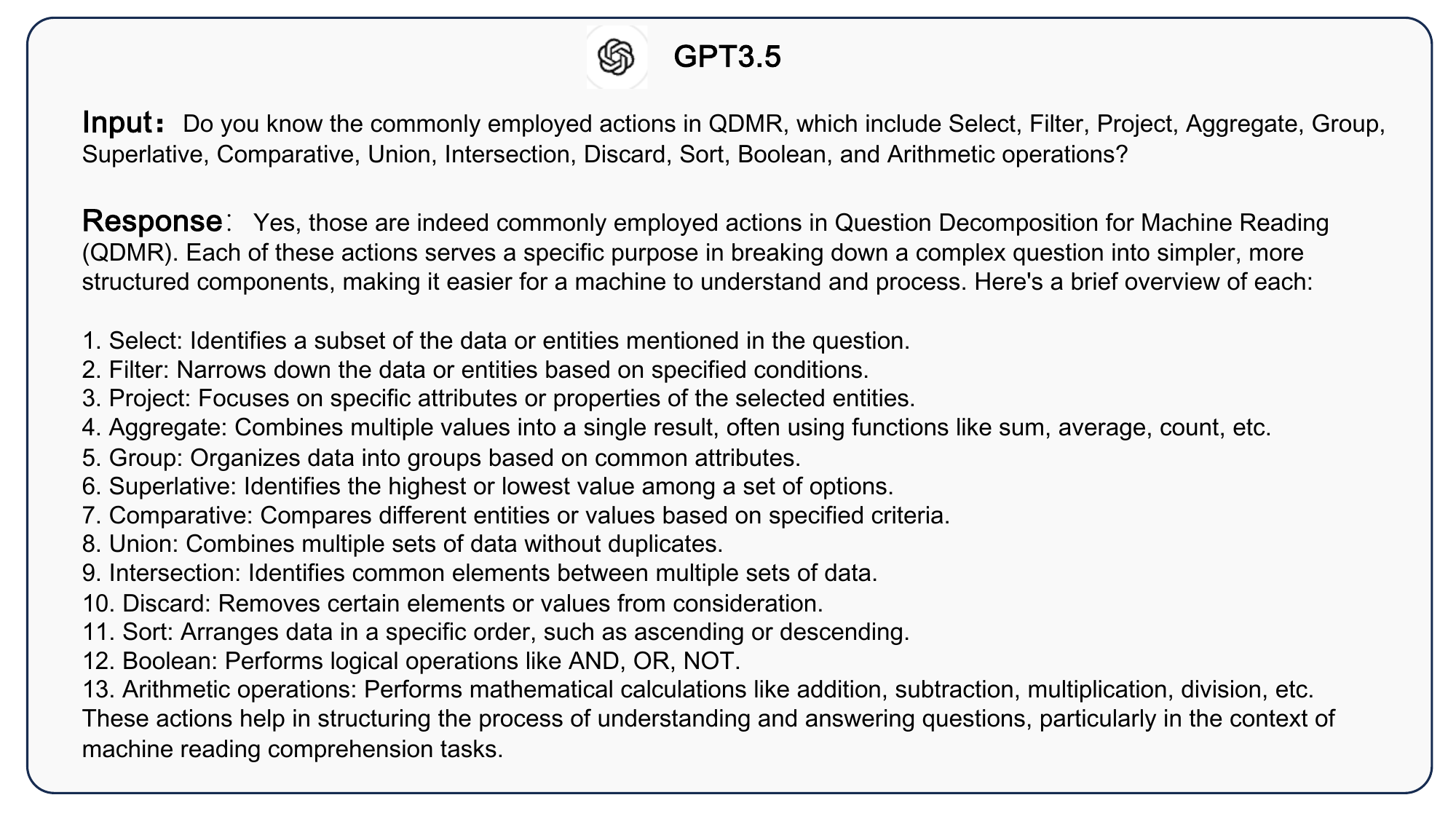}
\caption{The demonstration that the GPT-3.5 model includes the knowledge of actions within QDMR.}
\label{fig_gpt3.5_action}
\end{figure*}

In this section, we illustrate the prompt for executing the agent-style reasoning.
The complete prompt for AgentCOT is shown in Figure~\ref{fig_full_prompt}.
% The prompt designed for agent-style reasoning does not provide explicit action descriptions, as we have determined that the LLM already encompasses the knowledge of the action set in QDMR, as presented in XX and XX.
We can see that the prompt does not provide explicit action descriptions, as we have determined that the LLM already encompasses the knowledge of the action set in QDMR, as presented in Figure~\ref{fig_gpt3_action} and Figure~\ref{fig_gpt3.5_action}.

\begin{figure*}[ht]
\centering
\includegraphics[scale=0.47]{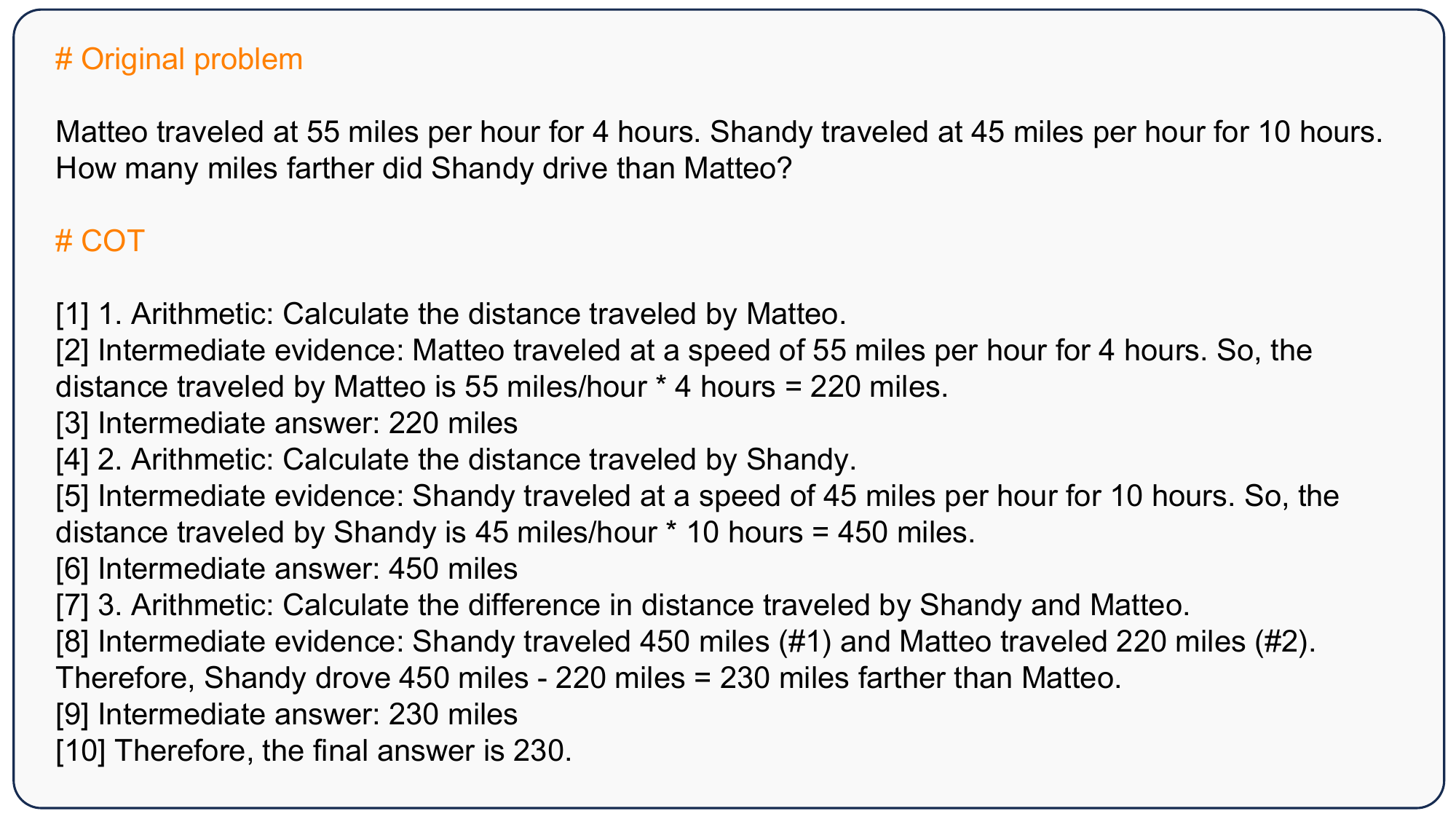}
\caption{An example of COT generated by AgentCOT. '[N]' is provided for readability purposes and is not part of the source sequence.}
\label{example_cot}
\end{figure*}

Here, we provide a detailed COT example generated by AgentCOT in Figure~\ref{example_cot}. 
When the original problem $Q$ is coming, AgentCOT first selects an action $a^0$ from a defined action set and delivers a detailed description $a_{des}^0$ of the selected action (line [1]). Then $Q+a^0+a_{des}^0$ replaced $Q$ is fed into AgentCOT to generate intermediate evidence $E_{inter}^0$ (line [2]) and intermediate result $R_{inter}^0$ (line [3]). At this point, AgentCOT has accomplished the first step in resolving Q. Next, AgentCOT responds to $Q+a^0+a_{des}^0+E_{inter}^0+R_{inter}^0$ and selects a new action $a^1$ with description $a_{des}^1$. AgentCOT iterates through the aforementioned process until the problem is solved.

In the implementation of AgentCOT, we encourage AgentCOT to generate the remaining complete inference process. For example, when AgentCOT first interacts with the original problem $Q$, it only needs to provide $a^0$ and $a_{des}^0$ (line [1])  but can also generate the remaining complete COT (line [1]-[10]). The complete COT is used for assessing whether AgentCOT's execution has terminated. If $a^i$, $a_{des}^i$, $E_{inter}^i$, $R_{inter}^i$ are the last step in complete COT, it indicates the problem-solving process has been finished.

% \setlength{\tabcolsep}{2.8mm}{
% \begin{table*}[]
% \centering
% \begin{tabular}{l}
% \toprule
% \\
% \toprule
% \end{tabular}
% \caption{}
% \label{tab:example}
% \end{table*}
% }

% Next, we present in-context learning demonstrations, displaying only one example per task for illustration.

\subsection{AgentCOT Enhancement}
\label{sec:agentcot_enhancement}
During self-evaluation decoding, AgentCOT assesses the current state by asking 'Is the current reasoning process reasonable?'.
This assessment process is based on LLM and occurred at step $i$ after generating \{$a^i$, $a_{des}^i$, $E_{inter}^i$, $R_{inter}^i$\}.

In the ensemble strategy, AgentCOT considers the action, the intermediate result, and the suggestive final result simultaneously. Taking the response when AgentCOT completes the first step in Figure~\ref{example_cot} as an example, the action is 'Arithmetic' (in line [1]), the intermediate result is '220 miles' (in line [3]), and the suggestive final result is '230' (in line [10]).
In the implementation process, we select the optimal current state based on a voting mechanism, with priority given to the suggestive final result, followed by the intermediate result, and finally the action. The self-evaluation decoding strategy is executed after the ensemble strategy.

% \subsection{AgentCOT with Core Information}
% \label{sec:6}
% Different from the approaches of producing chain-of-thought at once, agent-based AgentCOT
% requires multiple interactions to generate the final answer. This means that AgentCOT consumes more time and computational resources.
% We investigate the AgentCOT with core information in terms of performance and resources.
% We conduct experiments on CC and CSQA based on the model of \textit{gpt-3.5-turto} and results are shown in Table~\ref{tab:core_info}.

% By analyzing statistics, AgentCOT solving problems needs to consume 235 and 45 tokens on CSQA and CC on average respectively, while AgentCOT with core information only needs to manage 95 and 22 tokens. From the results, the iterative process based on core information does not result in a significant decrease in model performance.
% It is acceptable to make a slight decline in performance when largely reducing resource consumption. Therefore, we suggest inferring with core information, which is more efficient and resource-friendly. Discarding supporting evidence can be seen as a redundant memory information deletion strategy.

% \section{Prompt without Action Descriptions}

% The prompt designed for agent-style reasoning does not provide explicit action descriptions, as we have determined that the LLM already encompasses the knowledge of the action set in QDMR, as presented in XX and XX.

% 放在正文。
% There is no need to explicitly specify the actions since the LLM we use inherently possesses a deeper understanding of these actions.

% \lipsum[1-10]